\documentclass[12pt]{article}
\usepackage{amsmath}
\usepackage[margin=1in]{geometry} 
\usepackage{amssymb}
\usepackage{array}
\usepackage{epsfig}
\usepackage{graphicx}
\usepackage{xy}
\usepackage{tikz-cd}
\usepackage{algorithm}
\usepackage{algorithmic}
\usepackage{tcolorbox}
\usepackage{subcaption}
\usepackage{ulem}
\include{macros}
\title{Diffusion Maps : Using the Semigroup Property for Parameter Tuning}
\author{Shan Shan and Ingrid Daubechies}
\begin{document}
\maketitle
\begin{abstract}
Diffusion maps (DM) constitute a classic dimension reduction technique, for data lying on or close to a (relatively) low-dimensional manifold embedded in a much larger dimensional space. The DM procedure consists in constructing a spectral parametrization for the manifold from simulated random walks or diffusion paths on the data set. However, DM is hard to tune in practice. In particular, the task to set a diffusion time $t$ when constructing the diffusion kernel matrix is critical. We address this problem by using the semigroup property of the diffusion operator. We propose a semigroup criterion for picking the ``right" value for $t$. Experiments show that this principled approach is effective and robust.
\end{abstract}


\noindent 
\textbf{Keywords} 
\quad Diffusion maps 
\quad Diffusion operator
\quad Semigroup properties
\quad Manifold learning
\quad Dimension reduction.

\section{Introduction}
Diffusion maps (DM) \cite{coifman2006diffusion} are used in machine-learning to achieve dimension reduction for data that are assumed to be sampled from a lower-dimensional manifold within a higher-dimension\-al setting; they are related to other kernel eigenmap methods such as Laplacian eigenmaps \cite{belkin2003laplacian},  local  linear  embedding  \cite{roweis2000nonlinear},  Hessian  eigenmaps \cite{donoho2003hessian}  and local tangent space alignment \cite{zhang2007linear}.  

The basic idea is simple: diffusion on a manifold is governed by the semigroup generated by the manifold's Laplace-Beltrami operator; the spectral analysis of the diffusion operator thus provides information about the manifold that can be used to provide a lower-dimensional parametrization for the data that also removes ``noise'' from the data inconsistent with the manifold hypothesis.
One can (approximately) simulate a random walk or diffusion process on the (unknown) manifold by taking small steps within the data set according to probabilities estimated from the distances between data points. Indeed, if the data points provide a sufficiently dense sampling of the manifold, then their distances (measured in the high-dimensional ambient space)  within a close neighborhood of a fixed data point $P$ are close approximations to the distances between the corresponding points within the pull-back of the neighborhood to the tangent plane at $P$; the diffusion kernel on the manifold can be likewise approximated (near $P$) by that on the tangent plane at $P$. Since the diffusion kernel in a Euclidean space takes the same form (up to normalization) regardless of the dimension, one can thus simply use the distances in the ambient large-dimensional Euclidean space to generate a reasonable approximation to the manifold diffusion kernel for short diffusion times. 

More precisely, suppose we are given a set of points $\mathcal{D} = \{y_1, \dots, y_N \}$  residing in high-dimensional Euclidean space $\mathbb{R}^K$. To compute a low-dimensional representation of the data with DM, we first construct a diffusion kernel matrix $W^{(t)}$, 
\begin{align}
    W^{(t)}_{ij} = \exp\left(-\frac{d^2(y_i, y_j)}{t}\right). 
\end{align}
Here, $d(\cdot, \cdot)$ is a distance metric on $\mathcal{D}$, for example as defined by the $L_2$ norm $|| \cdot ||_2$ on $\mathbb{R}^K$, and $t$ is a {\it diffusion time} parameter chosen by the user. (This parameter is the ``short time'' from the hand-waving argument in the preceding paragraph.) Further operations are typically carried out (see below) to correct for possible local differences in e.g. sampling density within the data set. The resulting discrete matrix is interpreted as an approximation to the diffusion kernel (i.e. the kernel of the semigroup generated by the Laplace-Beltrami operator) on the manifold assumed to underlie the data. The entries $\left(\Psi_{\ell}\right)_i;\, \ell=1,\ldots,L,\, i=1,\ldots N$ (with $L \ll K$)
of this matrix's first few  eigenvectors $\Psi_{\ell}$, sometimes suitably weighted according to the eigenvalues $\lambda_{\ell}^t$, 
then provide an $L$-dimensional parametrization for the data points $y_i,\, i=1,\ldots N$.

This article is organized as follows. Section \ref{sec:background} gives a brief capsule description of the  algorithm for computing diffusion maps and its mathematical interpretation.  Section \ref{sec:difftime} discusses the sensitivity of the method to the choice for $t$, illustrates the difficulty in guesstimating the ``right'' $t$, and introduces the semigroup test as a criterion for determining a useful value for $t$;
we also give examples of using the semigroup test to the problems of finding optimal data embedding on synthetic data and real image data. Experiments show that this principled approach is effective and robust.

\section{Diffusion operator and Diffusion maps: a brief recap} \label{sec:background}

We present a very condensed summary; readers interested in more extensive discussion of Riemannian manifolds can consult e.g. \cite{boothby2003introduction}; 
the basics of Diffusion Maps can be found in \cite{coifman2005geometric}.

\subsection{Laplace Beltrami operator}

We begin by defining the Laplace-Beltrami operator applied to a scalar function $f: \mathcal{M} \rightarrow \mathbb{R}$ on a Riemannian manifold $\mathcal{M}$; for simplicity we shall assume $\mathcal{M}$ to be compact, without boundary. 

{\bf Definition} (Laplace-Beltrami) Let $\mathcal{M}$ be a Riemannian manifold with a metric $g$. The Laplace-Beltrami operator on $\mathcal{M}$ is defined by
\begin{align}
    \Delta_{\mathcal{M}} f(x) = - \text{Trace} \nabla^{\mathcal{M}} \nabla f(x),
\end{align}
where $\nabla^{\mathcal{M}}$ denotes the canonical {\it Levi-Civita connection} on ${\mathcal{M}}$ associated with $g$.

\noindent
(In the case where $\mathcal{M}$ is a compact manifold with a smooth boundary, one has to consider appropriate boundary conditions in order to define $\nabla_{\mathcal{M}}$ as a self-adjoint operator.) 

The spectrum of $\Delta_{\mathcal{M}}$ on a compact manifold 
$\mathcal{M}$ is discrete. Let the eigenvalues be $0 = \gamma_0 \leq \gamma_1 \leq \gamma_2 \leq \dots$ and let $f_i$ be the eigenfunction corresponding to eigenvalue $\gamma_i$. Then the eigenfunctions of the Laplace-Beltrami operator give rise to an embedding operation with certain optimality properties. 

An important observation (see e.g. \cite{belkin2008towards}),  is that $|| \nabla f ||$ provides us with an estimate of how far apart $f$ maps nearby points. Let $y_i, y_j \in \mathcal{M}$; then
\begin{align}
    |f(y_i) - f(y_j)| \leq \mbox{dist}_{\mathcal{M}}(y_i, y_j) || \nabla f || + o(\mbox{dist}_{\mathcal{M}}(y_i, y_j)),
\end{align}
indicating that points close together on the manifold are mapped by $f$ to values close together in $\mathbb{R}$.
The extent to which $f$ ``preserves'' locality can be measured by e.g.
\begin{align}
\label{eq:obj}
\int_{\mathcal{M}} || \nabla f(y) ||^2 d_{vol_{\mathcal{M}}}y;
\end{align}
minimizing this objective function
is equivalent to finding eigenfunctions of the Laplace Beltrami operator $\Delta_{\mathcal{M}}$, in the following sense. The eigenfunction $f_0$ of $\Delta_{\mathcal{M}}$ with the lowest value for (\ref{eq:obj}) is a constant function
on $\Delta_{\mathcal{M}}$ (for which $\nabla f(y)=0$ for all $y$), which is completely 
uninformative concerning localization of manifold points w.r.t. each other, since it maps all points to the same value in $\mathbb{R}$. The next eigenfunction 
$f_1$ provides an optimal embedding map to the real line (in the sense that it minimizes the integral over $\mathcal{M}$ of the averaged distortion bound (\ref{eq:obj})); similarly, the optimal embedding of the manifold in $\mathbb{R}^L$ is defined by
\begin{align}
    \mathbf{g} := \left(f_1(y), \dots, f_L(y)\right).
\end{align}

\subsection{Diffusion operator}
Although the first $L$ eigenvectors of the 
Laplace-Beltrami operator provide an informative embedding of $\mathcal{M}$ in $\mathbb{R}^L$, it can be
difficult to identify these eigenvectors with a reasonable degree of accuracy, starting from noisy samples of $\mathcal{M}$. For this reason, it may be useful, in order to determine (approximations to)
these eigenvectors, to work instead with the semigroup of operators $\{e^{-t\Delta_\mathcal{M}}\}_{t\geq 0}$.
The Laplace-Beltrami operator $\Delta_\mathcal{M}$ 
is the infinitesimal generator of $e^{-t\Delta_\mathcal{M}}$, i.e., 
\begin{align}
    \lim_{t \rightarrow 0} \frac{I - e^{-t\Delta_\mathcal{M}}}{t}f = -\Delta_\mathcal{M} f, 
\end{align}
whenever $f$ belongs to a suitable dense subset of $C({\mathcal{M}})$. 
The diffusion operators $\{e^{-t\Delta_\mathcal{M}} \}_{t>0}$ share the same eigenfunctions as  $\Delta_{\mathcal{M}}$, and the eigenvalues of $e^{-t\Delta_{\mathcal{M}}}$ are exactly the $e^{-\gamma_\ell t}$; in particular, they are bounded: $1 = e^{-\gamma_0 t} \geq e^{-\gamma_1 t} \geq e^{-\gamma_2 t} \geq \dots >0. $

The Laplace-Beltrami operator $\Delta_\mathcal{M}$ and the diffusion operator $e^{-t\Delta_\mathcal{M}}$ are also related through the {\em heat} (or {\em diffusion}) equation on the manifold. 

\textbf{Definition} (Heat Equation). Let $f: \mathcal{M} \rightarrow \mathbb{R}$ be the initial temperature distribution on a manifold $\mathcal{M}$ embedded in $\mathbb{R}^K$. The {\it heat equation} is the partial differential equation
\begin{align}
    \begin{split}
    &\frac{\partial u}{\partial t} + \Delta_{\mathcal{M}}u = 0, \\
    &u(x,0) = f(x).
    \end{split}
\end{align}
The solution of the heat equation is given by the diffusion operator $e^{-t\Delta_\mathcal{M}}$,
\begin{align}
    u(x,t) &= e^{-t\Delta_\mathcal{M}}f(x) \\
    &=  \int_{\mathcal{M}} h_t(x,y) f(y) d_{vol_{\mathcal{M}}}y.
    \label{eq:heatsolint}
\end{align}
In the integral from (\ref{eq:heatsolint}), $h_t$ is called the {\it heat kernel}. When $x,y$ are close to each other
and $t$ is small, $h_t$ can be approximated by the Gaussian 
\begin{align}
    h_t(x,y) = (4\pi t) ^{-\frac{d}{2}} e^{-\frac{||x- y||^2}{4t}},
    \label{eq:heatkernel}
\end{align}
where $d$ is the dimension of $\mathcal{M}$.

\subsection{Approximating the diffusion operator on a discrete dataset}

On discrete data samples of $N$ data objects in $K$ dimensions, interpreted as points of an unknown Riemannian manifold $\mathcal{M}$ embedded in $\mathbb{R}^K$, an approximation of the diffusion operator $e^{-t\Delta_\mathcal{M}}$ is built as follows.
First, define the $N$ by $N$ matrix $W$, 
\begin{align}
    W_{ij} = \begin{cases}
     e^{-\frac{||y_i - y_j||^2}{4t}} ~~&\mbox{if} ~||y_i - y_j|| < \epsilon \\
    0 ~~&\mbox{otherwise}
    \end{cases}
    \label{eq:gaussweight}
\end{align}
where $ \epsilon$ is picked in concordance with $t$; typically
$\epsilon$ is proportional to $(Ct)^{1/2}$ for some $C$ significantly larger than 1; the $ W_{ij}$ set to zero thus correspond to entries that would otherwise be so small that they
would not contribute much to the overall matrix, while setting them to zero alleviates the complexity of the algorithm. The action of the true diffusion kernel, acting as an integral operator on the constant function 1 on 
${\mathcal{M}}$, would produce the function 1 again; the discrete approximation $W$ typically does
not have the same effect on the all-ones vector approximating the function 1. Many of the approximation ingredients contribute to this shortcoming, such as setting some of the $W$-entries to zero, or (more importantly) local variations in the data manifold sampling, which result in some data points having more and/or closer neighbors than others and which also cause the summing (rather than integration) procedure to deviate from an optimal quadrature. To remedy the total effect of these shortcomings to some extent, one defines a diagonal
matrix $D = [D_{ii}]_{i=1}^N$, with entries given by $D_{ii} = \sum_j W_{ij}$; the matrix $D^{-1}W$ then does indeed map the all-ones vector to itself. (Note that this also neatly sidesteps the problem that we had no estimate for the dimension $d$ of $\mathcal{M}$, which would, in principle, have been necessary for the normalization of the gaussian approximation to $h_t$.) We then compute eigenvalues and eigenvectors for $D^{-1}W$, i.e.
\begin{align}
    D^{-1}W \mathbf{f_{\ell} = \lambda_{\ell} \mathbf{f}_{\ell}},
\end{align}
where we order the eigenvalues and eigenvectors so that
$1=\lambda_0 \geq \lambda_1 \geq \lambda_2 \geq \ldots~$. 
The $k$-dimensional embedding is then defined by
\begin{align}
    \mathbf{g}_i = (\mathbf{f}^{[i]}_{1},\dots, \mathbf{f}^{[i]}_{L}),
\end{align}
where $\mathbf{f}^{[i]}_{\ell}$ is the $i$-th entry in the $N$-dimensional $\ell$-th eigenvector.
\vspace*{.2 cm}

Figure 1 illustrates this embedding on a simple example. The dataset consists
of points residing on a helix wrapped around a torus, and shows the 2D
embeddings obtained by Laplacian eigenmaps and diffusion maps, comparing them 
with those from
PCA and MDS; Laplacian-based methods clearly do the better job recovering the intrinsic data structure -- in this example, PCA and MDS essentially give results similar to a linear projection onto a 2D-plane.

\begin{figure}[h]
 \label{fig:helix_nonlinear} 
 \centering   
 \includegraphics[height = 6 cm]{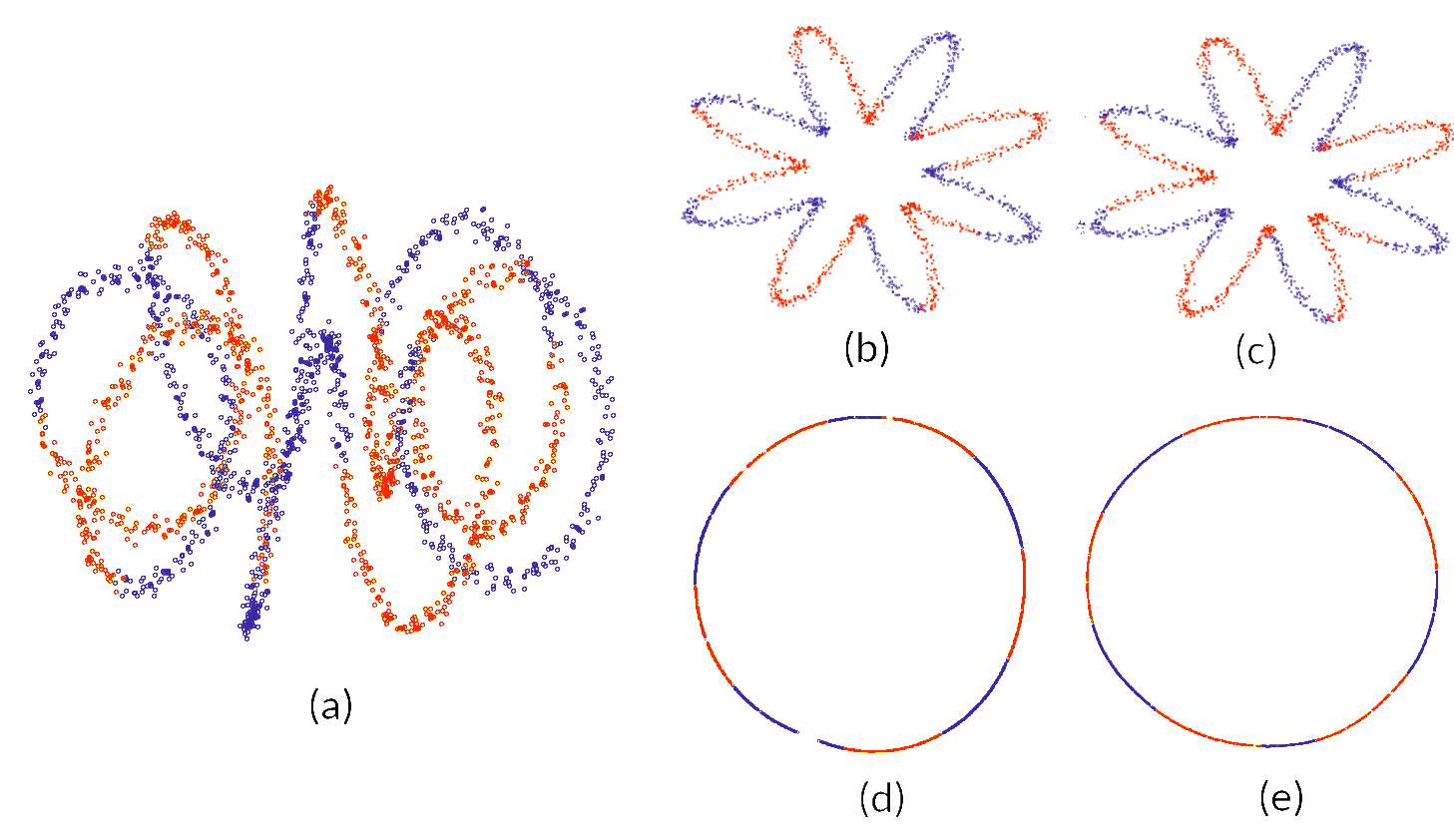}
\caption{\small{(a) Original data points, uniformly sampled (with some noise added) from a helicoidal curve wrapped around a torus in 3D. (We note that although we show the data embedded in a low-dimensional space in all our examples, the size of the ambient dimension has no impact on these methods, since they depend only on the distances between data points.)\\ Smaller panels:
2D embeddings of these data obtained via (b) PCA, (c) MDS, (d) Laplacian eigenmaps, and (e) Diffusion eigenmaps}}
\end{figure}
\vspace*{.2 cm}

\noindent
In a second example 
the data are sampled uniformly (without added noise) from a 2D ``Swiss roll" surface (a  rectangle rolled up so that it forms a spiral -- see Fig. 2(a)) embedded in 3D.
Figure 2
compares the embeddings produced by Laplacian eigenmaps and by Diffusion maps, showing that Diffusion maps introduce fewer deformations. 

\begin{figure}[h]
    \centering
    \includegraphics[height = 4cm]{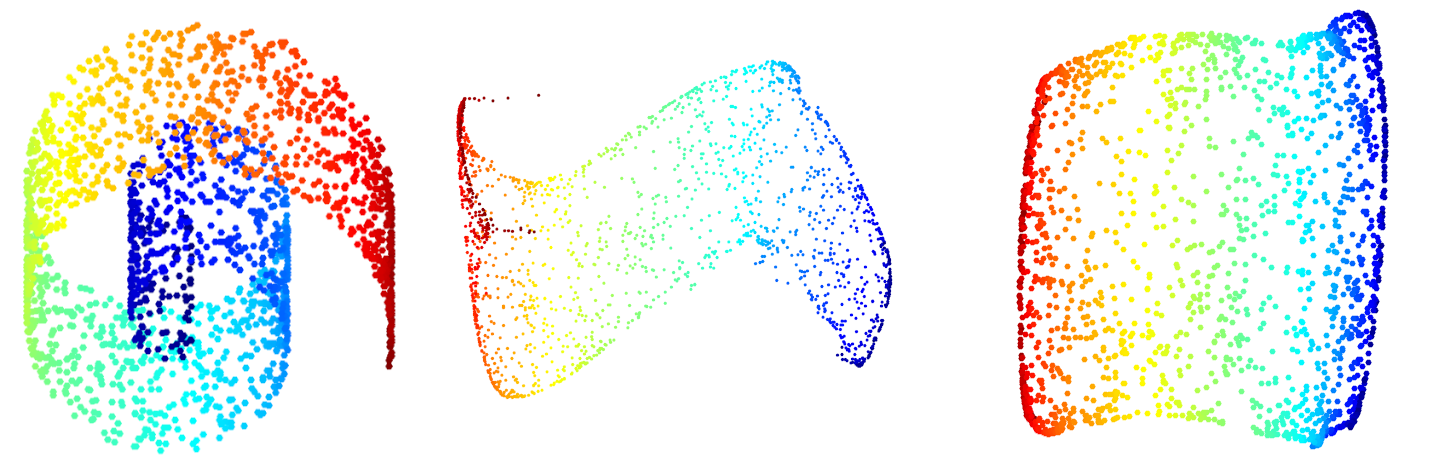}
    \caption{\small{Left: orginal Swiss roll data; Middle and Right: 2D Embeddings obtained via
Laplacian eigenmaps and Diffusion maps, respectively.}}
    \label{fig:swiss_nonlinear}
\end{figure}
\newpage

\noindent
{\bf Remarks}
$~$\\
1. The same {\em``normalization by left-multiplication by a diagonal matrix''} approach can be (and has been) used for the Laplace-Beltrami operator $\Delta_\mathcal{M}$ rather than $e^{-t\Delta_\mathcal{M}}$; approximating the differential operator by a matrix 
$L$ expressing a second-order difference, and then
``normalizing'' it by setting $D^{-1}L$, leads to the Laplacian eigenmaps 
proposed in \cite{belkin2008towards}. 

\noindent
2. Although the matrix $W$ is symmetric, the ``normalized'' version $D^{-1}W$
typically isn't. One can also consider instead the symmetrized version
$D^{-1/2}WD^{-1/2}$, which has the same eigenvalues as $D^{-1}W$; its eigenvectors are the vectors $D^{1/2}\mathbf{f}_{\ell}$.  

\noindent
3. In case the sampling density is known to be systematically not uniform over the manifold, it can be useful to 
introduce a correction for this in the matrix
construction. The paper \cite{coifman2006diffusion} describes in detail how
one can modify the construction to incorporate (an approximation to) the sampling
density. Depending on the value assigned to a tuning parameter $\alpha$, the
modification introduced in \cite{coifman2006diffusion} can be interpreted (when $\alpha =1 $) as introducing a Jacobian-like factor (so that, in the limit for finer and finer sampling, one recovers again the
standard diffusion semi-group and its Laplace-Beltrami generator), or 
(when $\alpha \neq 1$) as
an adjustment of the diffusion process itself, with a non-constant diffusivity; 
in the latter case, the approximation is linked to a different semigroup, the eigenvectors and eigenvalues of which encode again significant geometric information about the manifold, and can therefore again
be used for a lower-dimensional parametrization of the manifold. 
\vspace*{.4 cm}

\section{Setting the diffusion time $t$}
\label{sec:difftime}

\subsection{Sensitivity to the choice of $t$}

It is intuitively clear that the algorithm described earlier can work only in some window for $t$: if $t$ is chosen very large, then $W_{ij}$ will be different from zero even for
pairs $i,\,j$ for which the data points $y_i$ and $y_j$ are far from each other in the ambient space, although we expect that their 
Euclidean distance is not at all informative about their relative roles on the manifold 
$\mathcal{M}$. We argued in Section 2 that the construction of $\left(W_t\right)_{ij}$ (where we now explicitly denote the dependence of $W$ on $t$) was reasonable for $i,\,j$ where the distance 
$\|y_i-y_j\|$ was sufficiently small that $y_j$ could be viewed as close to the tangent plane to $\mathcal{M}$ at $y_i$; this means we should expect the method to work well only for $t$ below some threshold. This is also consistent with the proofs in \cite{coifman2005geometric} and \cite{coifman2006diffusion}: since those are proofs holding for $t \longrightarrow 0$, the similarity of the eigendecompositions of $W_t$ and the true diffusion operator $e^{-t \Delta_\mathcal{M}}$ on $\mathcal{M}$ can be expected only in the regime of small $t$. 

When $t$ is too small, the method faces a different problem: for $t< \min_{i\neq j}\|y_i-y_j\|^2$, 
$W$ reduces to the identity operator, and no useful embedding can be constructed. The problem persists for slightly larger values of $t$, where only a few $(i,j)$  emerge above the threshold. In a certain sense, the
diffusion time is then too short for the diffusion process to consistently bridge the distance between sample points on $\mathcal{M}$. Ideally, one would like that for each $i$,  $\left(W_t\right)_{ij}\neq 0$
for several $j \neq i$. (One would also like the number of such ``useful'' neighbors not to vary by orders of magnitude over the dataset. This is possible only if the sampling is fairly uniform. It is when the spatial distribution of the points in the dataset varies so much that no single parameter setting in the definition of $W_t$ allows for the number 
$\#\{j \neq i; \, \left(W_t\right)_{ij}\neq 0\}$ to be at least (say) 10 for all $i$ without getting into the several 100s for other $i$, that it is necessary to adapt the simple diffusion operator $W_t$, e.g. using the methods in \cite{coifman2006diffusion}.)

Finding the ``right'' choice for $t$, in the happy medium between the two extreme regimes, can be tricky: as illustrated in Figures 3 and 4 below, 
different choices for $t$ can lead to very different outcomes for the same data. 
\vspace*{-.1 cm}

\begin{figure}[h]
    \begin{minipage}{.5 \textwidth}
    \hspace*{-.04 \textwidth}
    \includegraphics[width =\textwidth]{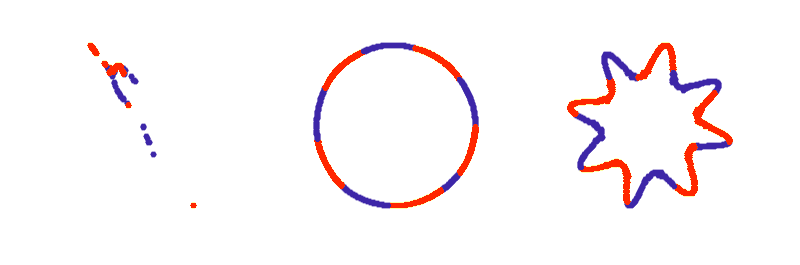}
    \end{minipage}
    \begin{minipage}{.48 \textwidth}
    \hspace*{.02 \textwidth}
    \caption{\small{Three embeddings of the helicoidal data from Figure
    1 obtained via Diffusion maps, for different choices for the 
    parameter $t$: $t_1$ (left), $t_2$ (middle) and $t_3$ (right). The choice  $t=t_2$ was adopted for Fig.1(e); the choices 
    $t=t_1=t_2/16$ and $t=t_3=4t_2$ are clearly suboptimal.}}
    \end{minipage}
\end{figure}

\begin{figure}[h]
    \includegraphics[width = \linewidth]{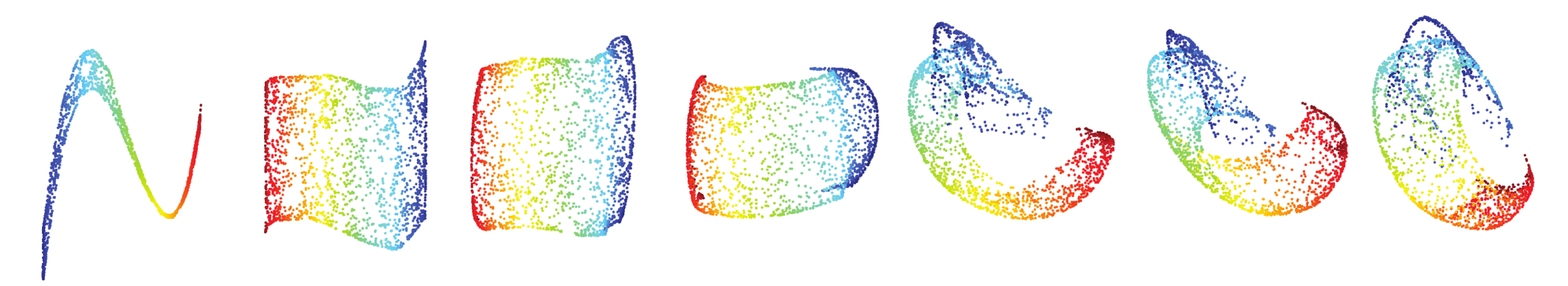}
    \caption{\small{Several embeddings of the Swiss roll data from Fig. 2 obtained by the Diffusion map
    procedure of Section 2; the
    only difference lies in the choice of the parameter $t$, increasing from left to right in the figure by a factor 2 each time.}}
    \label{fig:semigroup_nf}
\end{figure}

\subsection{Semigroup test}
In practical applications of Diffusion maps, it can take 
quite a bit of trial and error to find a ``right'' value for
$t$. 
Our goal here is to describe a simple 
robust guiding strategy to reduce this guesswork, which
finds a near-optimal value for $t$ in many situations in
which we have tested it.

The diffusion operators $\{e^{-t \Delta_\mathcal{M}} \}_{t>0}$ form a strongly continuous semigroup;
i.e. the $T_t:=e^{-t \Delta_\mathcal{M}} $ satisfy
\begin{align}
    T_{t_1+t_2} = T_{t_1}T_{t_2}\,, ~ \mbox{ and } s\mbox{-}\!\lim_{t\rightarrow 0}T_t = \mbox{Id}~.
\end{align}
It follows that the matrices $D_t^{-1}W_t$, used to define diffusion maps, or their symmetrized versions $K_t:=D_t^{-1/2}W_tD_t^{-1/2}$, can be approximate,
discretized versions of the diffusion operators on $\mathcal{M}$ only when they likewise
(approximately) satisfy the semigroup property.

We use this insight to formulate a criterion to pick an ``optimal''
$t$. 
In the regime where the $K_t$-operators are reasonable approximations of the
semigroup $\{e^{-t\Delta_\mathcal{M}}\}_{t\geq 0}$, 
 $\left( K_{t} \right)^2$ should be close to $K_{2t}$. 
This motivates the definition of 
the {\it semi-group error} $\mbox{(SGE)}(t)$,
\begin{align} 
\text{SGE}(t) := \| \left( K_{t} \right)^2 - K_{2t}\|. 
\end{align}
The norm used here is the operator norm; the operators we consider are (expected to be) positive, with eigenvalues between 0 and 1, and the range for SGE$(t)$ is between 0 and 1. 

In practice, 
we begin with initializing a wide range of discrete values for $t$, 
i.e. we pick a set $T:=\{t_m\,;\, m=1,\ldots M\}$.  
For each  $t_m$ in $T$, we construct the diffusion matrices $\left( K_{t_m} \right)^2$ and $K_{2t_m}$, and we compute  
$\mbox{(SGE)}({t_m})$.

Figure 5 below plots
the semi-group error SGE$(t)$ for different values of $t$ for the Swiss roll example of Figures 2 and 4; 
SGE$(t)$ reaches its lowest value
for the choice of $t$ where the 2D-embedding is closest to a rectangle, which we know to be the ground truth in this case. 

\begin{figure}[h]
    \centering
    \includegraphics[height = 6cm]{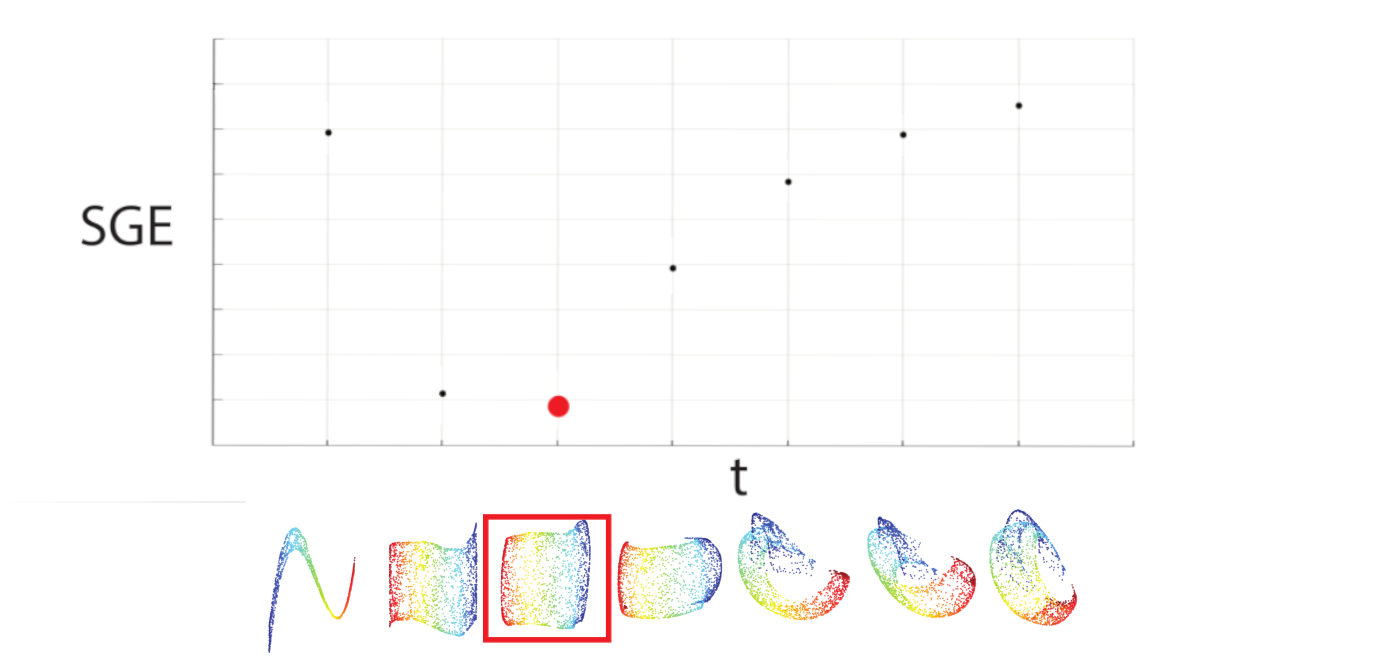}
    \caption{\small{Semi-group error SGE for the Swiss roll data, for the values of $t$ illustrated in Figure 3. 
    The red dot is the optimal value of $t$, and corresponds to the embedding that visually best reflects the ground truth. The scale for $t$ is logarithmic; each of the successive $t_m$ (at the tick marks) is larger by factor 2 than the previous one, $t_{m+1}=2\,t_m$.}}
    \label{fig:semigroup}
\end{figure}

In Figure 6, below, we revisit the three embeddings shown in Figure 3, next to the semi-group error plot for this dataset, and we observe that the visually optimal embedding (which is also the most accurate version of the ground truth for this manifold $\mathcal{M}$) corresponds again to
the value of $t$ with the smallest SGE in the regime of small diffusion times. 

\begin{figure}[h]
    \includegraphics[width = .9\textwidth]{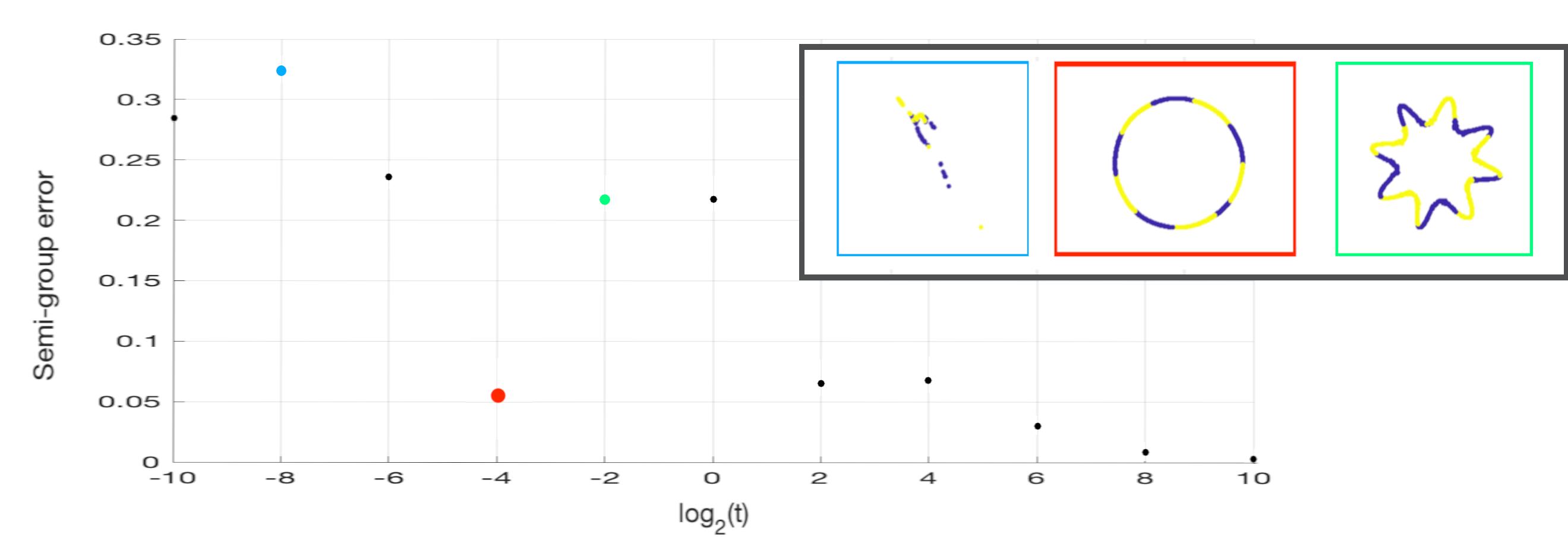}
    \caption{\small{Semi-group error plot for a wide range of candidate values for $t$ -- the ratio $t_{\max}/t_{\min}$ equals $2^{20}$ ! --  for the dataset illustrated in Figures 1 and 3. Note that for very large $t$, the SGE estimate becomes small again -- see discussion in the text.  } }
    \label{fig:circle_uniform}
\end{figure}

As shown in Figure 6, the values of SGE$(t)$ behave qualitatively as we would have expected, based on the intuition explained at the start of this section: when $t$ is very small, the semigroup behavior of the
$K_t$ hasn't ``kicked in'' yet, because the numerical diffusion's range is too short, and this is reflected by larger values for SGE$(t)$. (We recall that the range of SGE values is between 0 and 1; values exceeding .3 are indeed ``large''.) 
As $t$ increases, SGE$(t)$ drops to lower values, to start 
increasing again after a minimum SGE-value not too far above 0. (These small values are maintained in an interval for $t$, as may not be evident from Figure 6, in which the successive values of $t$ increase by a factor 4, $t_{m+1}=4\,t_m$; detailed behavior in the neighborhood of each $t_m$ is not apparent from this figure.) We interpret this increase as the influence of ambient-space geometry (such as the toroidal winding in this example), once the numerical diffusion is no longer ``following'' the manifold $\mathcal{M}$; because even noisy sampling from $\mathcal{M}$ translates to very non-uniform sampling in the higher-dimensional ambient space, the $K_t$ are less close to following a semi-group behavior. When $t$ increases further, the value
of SGE$(t)$ starts decreasing again: once the reach of the numerical diffusion is sufficiently large that the ``sources'' on or near $\mathcal{M}$ all ``act'' as one diffuse blob, and the geometry of $\mathcal{M}$ has been obscured, the semigroup nature of the ambient-space diffusion takes over. Although SGE$(t)$ is small again, one cannot use these diffusion maps to generate an informative low-dimensional embedding of $\mathcal{M}$ for $t$ in this range. 

We next turn to a few examples with non-uniform sampling. In this case we compensate for the change in sampling density by using the techniques described
in \cite{coifman2006diffusion}, using an integral kernel
$A_{t,\alpha}$ obtained by a ``renormalization'' of $W_t$. 
Regardless of the parameter setting for $\alpha$ that gives the best results (which depends on the type of non-uniformity), 
the basic intuition underlying the 
method remains the same: the spectral analysis, used to construct 
a low-dimensional embedding of $\mathcal{M}$, is predicated on the $A_{t,\alpha}$
approximating the kernels of a semigroup of operators. One can thus again use the SGE to determine optimal choices of $t$. 
Figure 7 shows the results for a non-uniformly sampled circle. 

\begin{figure}[h]
     \centering
    \includegraphics[width=\textwidth]{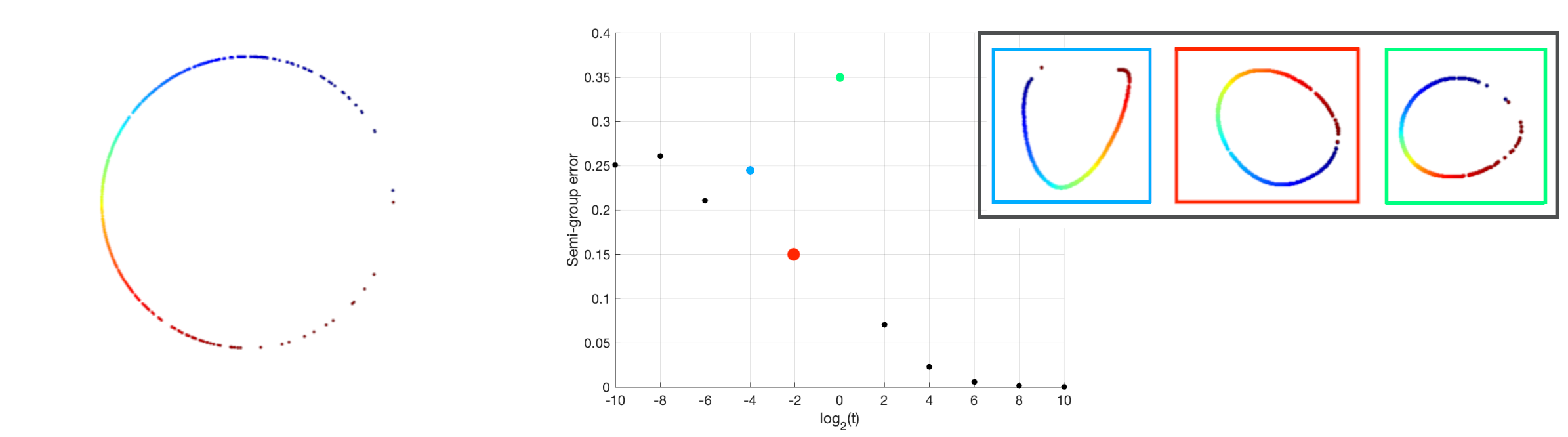}
     \caption{\small{Left: 512 non-uniformly sampled points on a circle; Middle: SGE$(t)$ for a wide range of $t$ values, with the useful-range optimal $t$ marked in red; Upper right inset: the embeddings obtained for the optimal $t=1/4$ and for $t=1/16$ (left) and $t=1$ (right). In this example, we set $\alpha =2$ to determine the adapted kernels $A_{t,\alpha}$. (In this case, there is hardly any dimension reduction, since both the original and final circle are depicted on a 2D plane; this is a toy example, after all. It may be worth noting that the parametrization by Diffusion maps would have been identical had the data been embedded in a random not-coordinate-aligned plane in a much higher dimensional space.)}}
    \label{fig:circle_nonuniform}
\end{figure}

\begin{figure}[h]
     \centering
    \includegraphics[width=\textwidth]{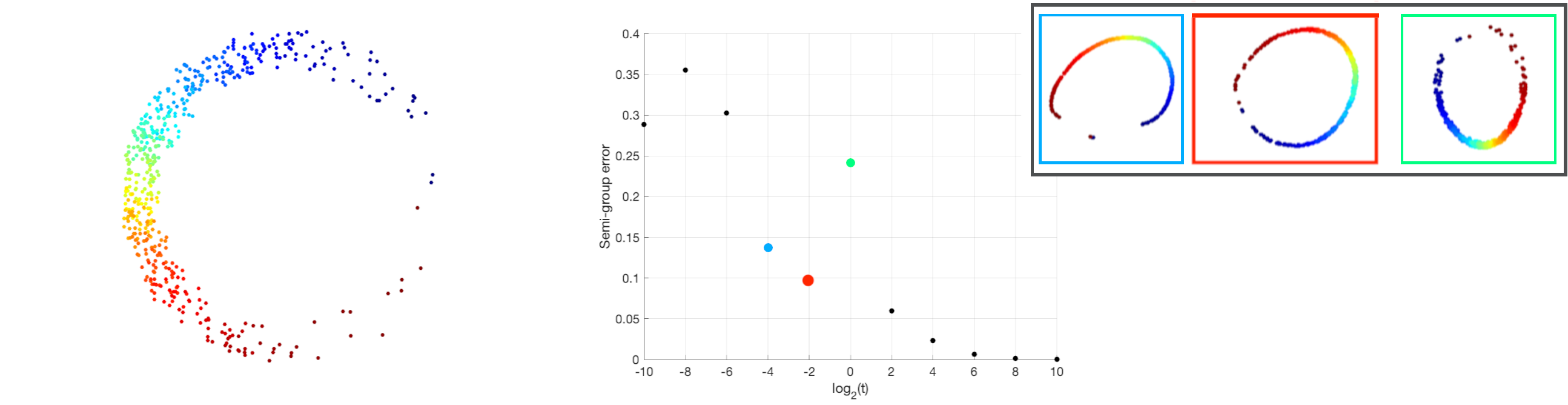}
     \caption{\small{Left: 512 non-uniformly sampled points on a circle, now with noise added; Middle: SGE$(t)$ for a wide range of $t$ values, with the useful-range optimal $t$ marked in red; Upper right inset: the embeddings obtained for the optimal $t=1/4$, as well as for $t=1/16$ (left) and $t=1$ (right). In this case, we have again set $\alpha =2$ to determine the adapted kernels $A_{t,\alpha}$.}}
    \label{fig:circle_nonuniform} 
\end{figure}

To illustrate the robustness of our the SGE test, we examine this dataset
again after noise has been added. The results are shown in Figure 8.

After the simulated toy data examples, we conclude with one example of real data. 

The dataset is part of the
extended Yale Face Database B \cite{georghiades2001few}; it consists of 64 images (192 pixels by 168 pixels) of the same human face, in the same pose, under different illumination conditions: the light source is moved around in two different directions. The intrinsic dimensionality of this collection of images is therefore expected to be 2, although the images themselves are objects in a much higher-dimensional space. We applied Diffusion maps,
coupled with the semigroup-error tuning strategy described above, to this collection; the results are shown in Figure 9. By its very nature, the dataset in this example is noisy, since all photographs (as opposed to images generated by computer graphics) are inherently noisy, but we don't have an explicit chracterization of this noise. To illustrate robustness of our analysis and semigroup criterion to noise, we resort to a common strategy in image analysis: we revisit the dataset in Figure 10, after extra noise has been added independently to each of the 64 images.

To add noise to the datapoints, from Figure 9 to Figure 10, we proceeded as follows. For each of the $192\times 168$ pixels in each of the 64 images, we generated a random integer $I$ uniformly in $[-100,100]$; we then replaced the pixel value $P$ by $P+I$ if $0<P+I<255$, by 0 if $P+I<0$ or by 255 if $P+I>255$. An example of one of the face photographs, before and after adding noise, is shown in Figure 11 below. 

Despite the severity of the noise, we observe that the Diffusion Map analysis, combined with the semigroup tuning strategy, is remarkably robust: the same $t_{\mbox{\tiny{opt}}}$ is selected in both cases, and the corresponding 2D
embeddings are very similar (up to an inversion of the axis in one of the 2 
variables), as illustrated by Figure 12 below. 

\begin{figure}[t]
    \includegraphics[width= \textwidth]{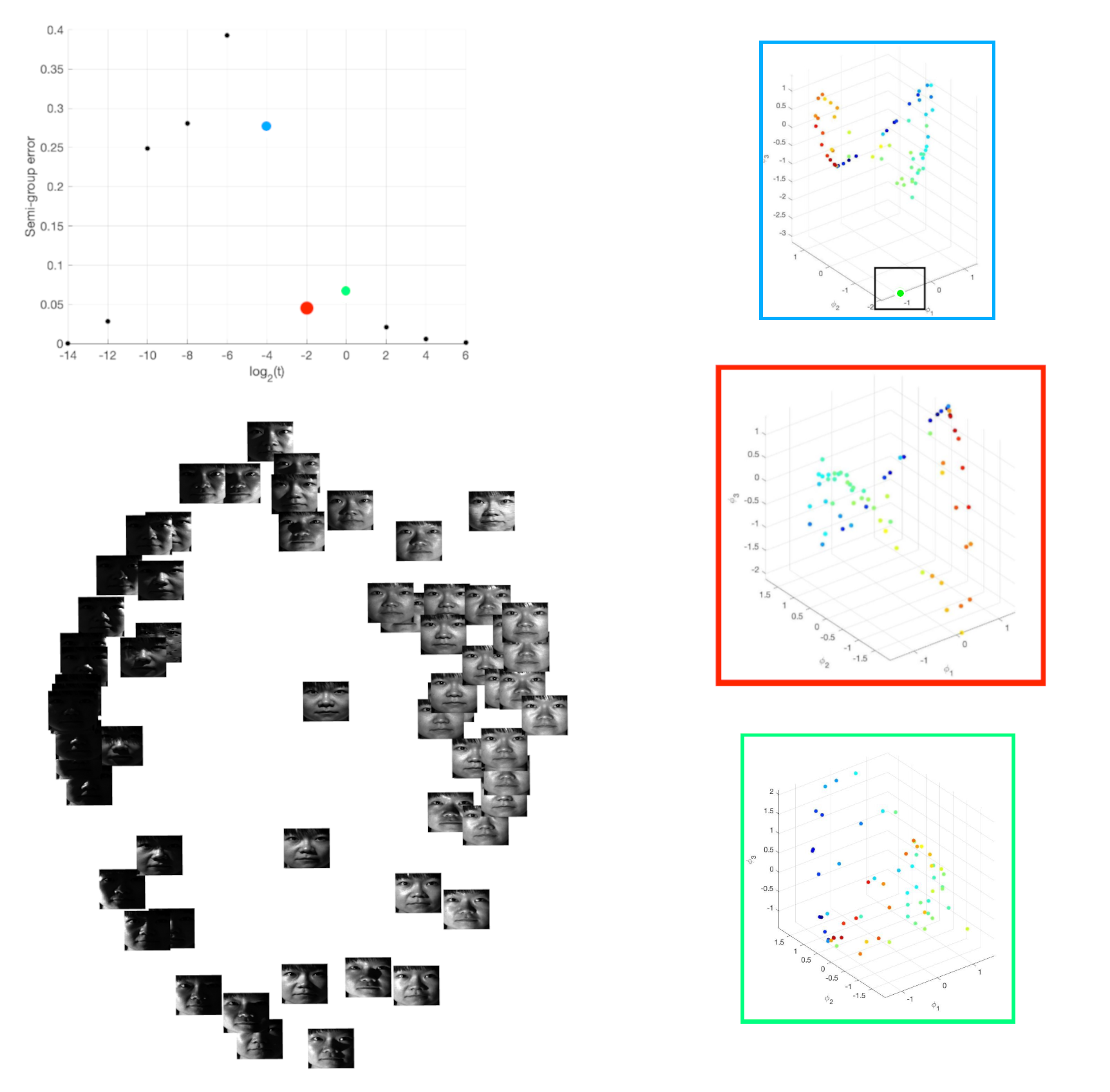}
     \caption{\small{DM with the semigroup tuning strategy applied to the Yale Face dataset, in which each data point is an image of $192 \times 168$ pixels. Top left: SGE$(t)$ for a wide range of $t$. The colored boxes above show the embeddings using the first 3 non-trivial eigenvectors for the optimal choice
     $t_{\mbox{\tiny{opt}}}=1/4$ and its two SGE-plot neighbors 
     $t_{-4}=t_{\mbox{\tiny{opt}}}/4$ and $t_0=4\,t_{\mbox{\tiny{opt}}}$. Although the embedding for $t_{-4}$ looks comparable to that for 
     $t_{\mbox{\tiny{opt}}}$ at first sight, the embedding shows that one data point is not integrated well with the rest (boxed green datapoint near one of the axes); for $t_{0}$ the structure of the dataset is much less well-defined than for $t_{\mbox{\tiny{opt}}}$. Left: 2-dimensional DM embedding for $t_{\mbox{\tiny{opt}}}$, with the datapoints indicated by thumbprints of the 
     images, indicating that the DM parametrization captured the illumination degrees of freedom in the dataset. (In this case, we set $\alpha = 1.5$ to determine the adapted kernels.)}}
    \label{fig:yale_face}
\end{figure}

\begin{figure}[h]
    \includegraphics[width=\textwidth]{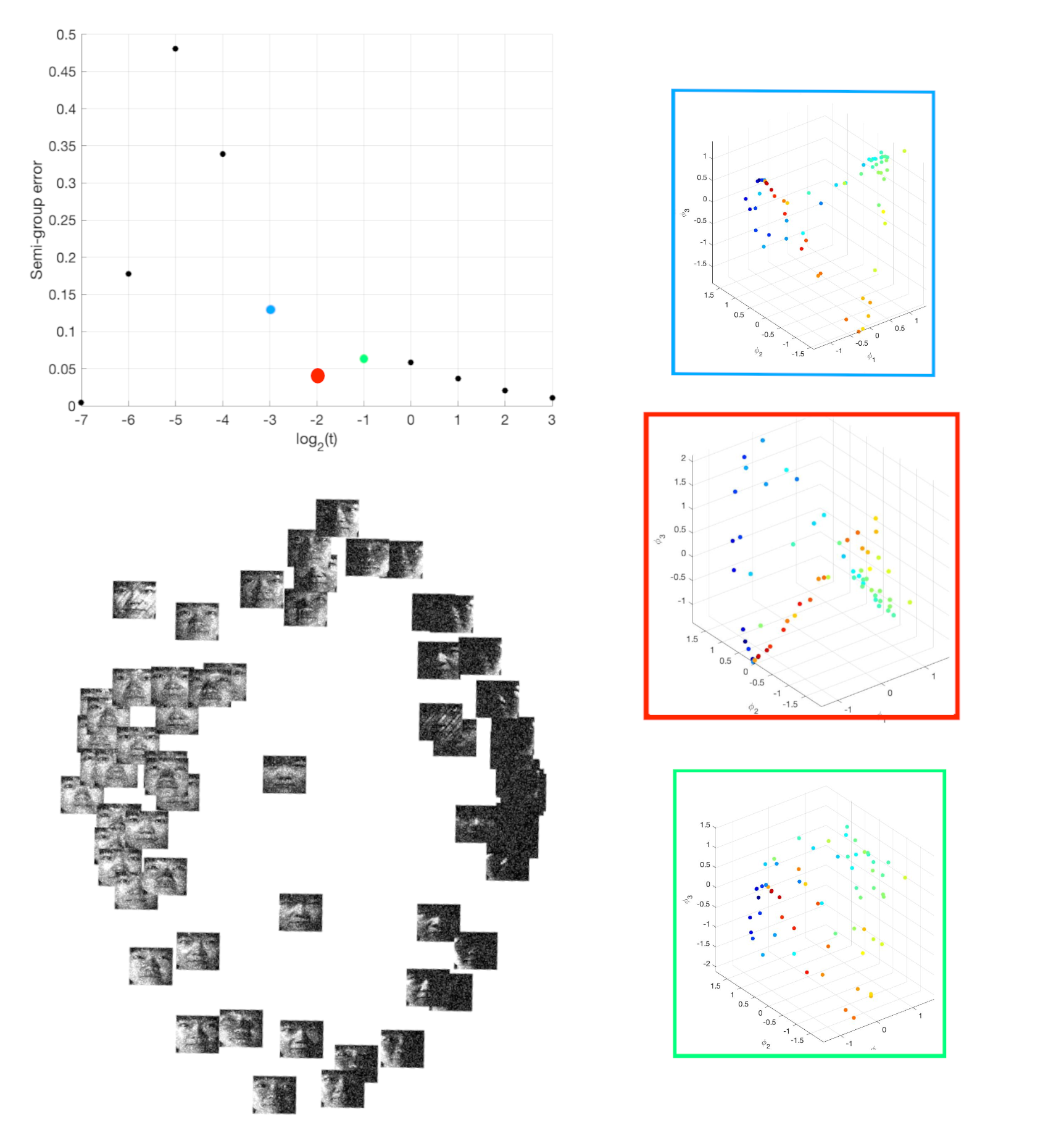}
     \caption{\small{DM with the semigroup tuning strategy applied to the same Yale Face dataset as in Figure 9, after additional noise was added to each image of $192 \times 168$ pixels, i.e. to each data point. (The nature of the noise is explained below and illustrated in Figure 11.) Top left: SGE$(t)$ for a wide range of $t$. The colored boxes above show the embeddings using the first 3 non-trivial eigenvectors for the optimal choice
     $t_{\mbox{\tiny{opt}}}=1/4$ and its two SGE-plot neighbors 
     $t_{-4}=t_{\mbox{\tiny{opt}}}/4$ and $t_0=4\,t_{\mbox{\tiny{opt}}}$.  Left: 2-dimensional DM embedding for $t_{\mbox{\tiny{opt}}}$, with the datapoints indicated by thumbprints of the 
     images. (We again set $\alpha = 1.5$ to determine the adapted kernels.)}}
    \label{fig:yale_face_noise}
\end{figure}

\begin{figure}[h]
\centering
\includegraphics[width=.6\textwidth]{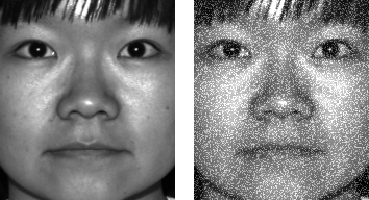}
\caption{\small{Left: one of the 64 images from the Yale Face dataset used in the analysis in Figure 9; Right: the noisy version of this image obtained by adding, pixelwise, random integers picked uniformly and independently in [-100,100] to the gray value of the pixel, and rounding so the result is in [0,225].}}
\end{figure}

\begin{figure}[h]
\centering
\includegraphics[width=\textwidth]{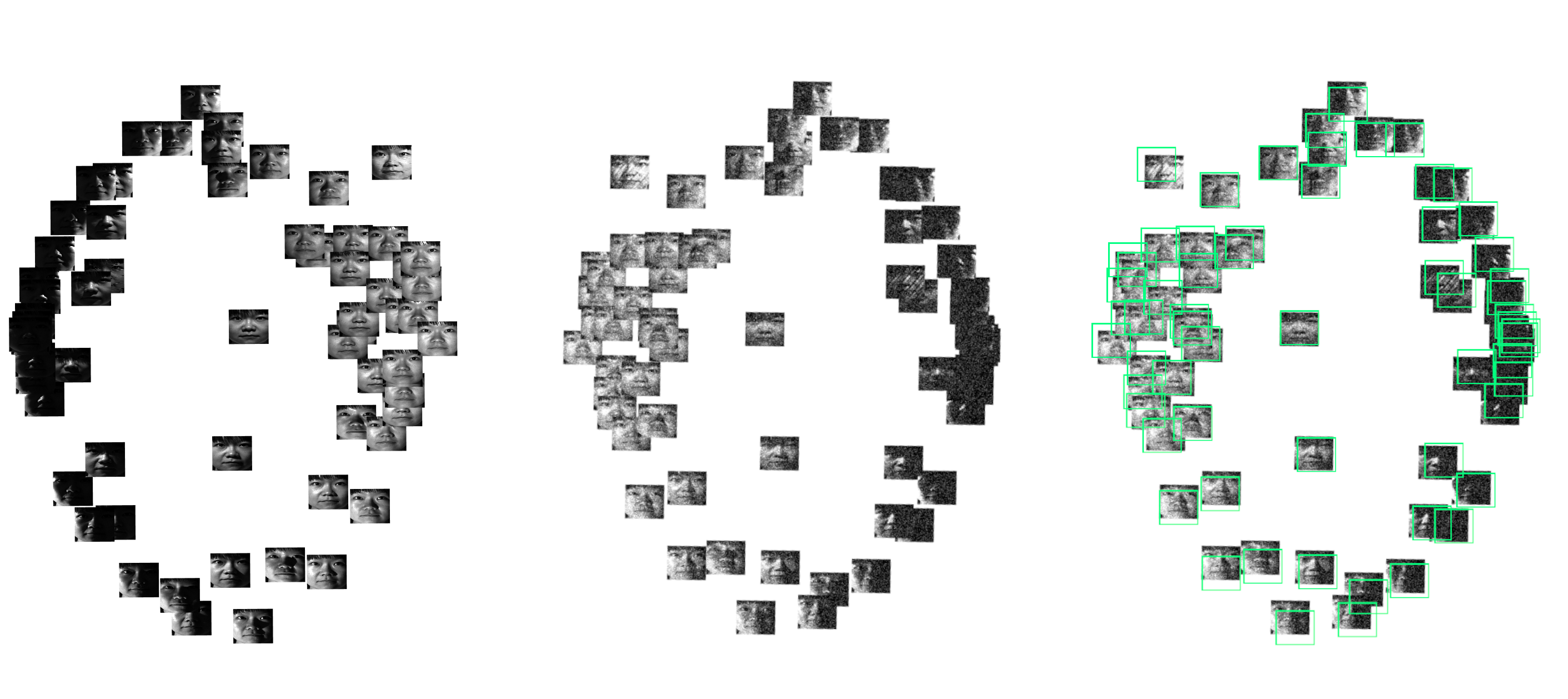}
\caption{\small{Left and Middle: the 2D embeddings from Figures 9 and 10, respectively, for the 64-face Yale Face dataset without and with the added noise illustrated in Figure 11. Up to a change of sign in the horizontal axis, the two embeddings are remarkably similar, as illustrated by the figure on the Right, which superimposes onto the Middle embedding a ``skeleton'' of the other embedding, indicating with a green rectangle the mirrored position of each thumbprint from the Left embedding. Closer scrutiny shows that thumbprints in close geometric proximity in this comparison picture do indeed correspond to the same face picture.}}
\end{figure}

\section{Conclusion}

Although Diffusion maps have shown to be a powerful tool to explore datasets embedded in high dimensions that are suspected to have interesting geometric structure on a much lower-dimensional scale \cite{liu2009learning}\cite{van2018recovering}\cite{moon2019visualizing}, 
determining the ``right'' value for the
diffusion parameter $t$ has been found to be tricky. Picking $t$ so that
it minimizes the Semi-Group Error is computationally easy, makes sense from a theoretical point of view, and gives good results in practice. 

\bibliographystyle{abbrv}
\bibliography{ref} 
\end{document}